\documentclass[twocolumn,11pt]{article} 
\usepackage[utf8]{inputenc} 
\usepackage{lipsum} 
\usepackage{svg}
\usepackage{amsfonts}
\usepackage{amsmath}
\usepackage{natbib}
\usepackage{graphicx}

\usepackage{subcaption}
\usepackage{titlesec} 
\usepackage{orcidlink}
\usepackage{fancyhdr} 
\usepackage{xcolor} 
\usepackage{hyperref} 
\usepackage{authblk}
 \usepackage{float}
\hypersetup{
    colorlinks=true,
    linkcolor=blue,
    citecolor=red,
    urlcolor=green,
    pdftitle={A Hybrid Quantum-Classical Neural Network for Interpretable Brain Tumor Classification}
}

\fancypagestyle{mystyle}{
    \fancyhf{}
    \fancyhead[L]{HQCM-EBTC: A Hybrid Quantum-Classical Model for Explainable Brain Tumor Classification}
    \fancyfoot[C]{\thepage}
}
\title{HQCM-EBTC: A Hybrid Quantum-Classical Model for Explainable Brain Tumor Classification}

\author[1]{Marwan Ait Haddou \orcidlink{0009-0008-1734-1721}. \footnote{aithaddou.marwan@edu.uca.ac.ma}}

\author[1]{Mohamed Bennai \orcidlink{0000-0002-7364-5171}.\footnote{mohamed.bennai@univh2c.ma}}

\affil[1]{Quantum Physics and Spintronic Team, LPMC, Faculty of Sciences Ben M’sick,
Hassan II University of Casablanca, Morocco}

\begin{document}

\onecolumn
\maketitle
\begin{abstract}
    This study investigates the efficacy of a hybrid quantum-classical model, denoted as HQCM-EBTC, for the automated classification of brain tumors, comparing its performance against a classical counterpart. A comprehensive dataset comprising 7,576 magnetic resonance imaging (MRI) images, encompassing normal brain structures, meningioma, glioma, and pituitary tumors, was employed. The HQCM-EBTC model integrates a quantum processing layer with 5 qubits per circuit, a circuit depth of 2, and 5 parallel circuits, trained via the AdamW optimizer with a composite loss function that combines cross-entropy and attention consistency losses.

    The results demonstrate that HQCM-EBTC significantly outperforms the classical model, achieving an overall classification accuracy of 96.48\%  compared to 86.72\% . The quantum-enhanced model exhibits superior precision, recall, and F1-scores across all tumor classes, particularly in glioma classification. t-SNE visualizations reveal enhanced feature separability within the quantum processing layer, leading to more distinct decision boundaries. Confusion matrix analysis further substantiates a reduction in misclassification rates with HQCM-EBTC. Moreover, attention map analysis, quantified using the Jaccard Index, indicates that HQCM-EBTC produces more localized and accurate tumor region activations, especially at higher confidence thresholds.

    These findings underscore the potential of quantum-enhanced models to improve brain tumor classification accuracy and localization, offering promising advancements for clinical diagnostic applications. The demonstrated ability of HQCM-EBTC to achieve higher accuracy and more precise tumor localization suggests a significant step forward in applying quantum computing to medical imaging analysis. 
\end{abstract}

\twocolumn
\section{INTRODUCTION} \label{introduction}

Brain tumors represent abnormal masses of tissue that proliferate uncontrollably within the brain, lacking normal physiological function \citep{kolb1985fundamentals}. These neoplasms encompass a diverse group of entities characterized by erratic cell growth. Although brain tumors occur less frequently than other cancers \citep{vienne2019environmental}, they pose a severe threat---especially in children---where they rank among the leading causes of tumor-related mortality. Recent data indicate that the incidence rates of brain tumors have surged, which has intensified the urgency to address this growing public health concern.

In the United States, brain tumors stand as the 10th leading cause of death, affecting an estimated 700,000 individuals, with approximately 80\% of cases classified as benign and 20\% as malignant \citep{rahman2022internet}. The American Cancer Society projects that 24,820 malignant brain and spinal cord tumors (14,040 in males and 10,780 in females) will be diagnosed in 2025, resulting in an estimated 18,330 deaths \citep{american2025statistics}. Globally, brain tumors continue to cause substantial cancer-related mortality among both pediatric and adult populations \citep{ayadi2021deep}.

Medical imaging, and MRI in particular, plays a pivotal role in detecting and diagnosing brain tumors. Institutions such as the University Hospital of Geneva generate between 12,000 and 15,000 images daily \citep{al2021mri}, yet the manual evaluation of these images remains labor-intensive, costly, and prone to human error. Consequently, there is a compelling need for automated diagnostic solutions that streamline image analysis and improve accuracy.

Deep learning (DL), especially convolutional neural networks (CNNs), has achieved remarkable success in tumor classification by extracting and learning complex imaging features. Researchers have harnessed pre-trained CNN models to boost diagnostic accuracy; however, classical DL approaches still encounter significant hurdles, such as high computational costs, limited scalability, and difficulties generalizing to high-dimensional medical datasets \citep{litjens2017survey}.

Quantum machine learning (QML) has recently emerged as a promising paradigm that overcomes some of the inherent limitations of classical machine learning. By exploiting quantum principles such as superposition, entanglement, and quantum parallelism, QML enables more efficient data processing and the capture of intricate patterns that are difficult for classical models to discern \citep{schuld2019quantum, biamonte2017quantum}. Quantum models---such as quantum support vector machines (QSVMs) \citep{rebentrost2014quantum}, quantum neural networks (QNNs) \citep{farhi2018classification}, and quantum convolutional neural networks (QCNNs) \citep{cong2019quantum}---have demonstrated improvements in both diagnostic accuracy and computational efficiency. In particular, hybrid quantum-classical architectures have shown promise in tasks such as feature extraction, pattern recognition, and classification \citep{amin2023detection, kumar2024brain}.

In addition to enhancing classification accuracy, ensuring interpretability in medical AI models is imperative for fostering clinician trust, patient safety, and regulatory compliance. Black-box models that lack transparency are often met with skepticism by clinicians, who require insight into how automated decisions are derived before incorporating them into clinical practice \citep{holzinger2019causability}. Moreover, interpretability is vital in mitigating AI-driven diagnostic errors, as clear explanations facilitate alignment with established medical knowledge and aid in the detection of biases stemming from imbalanced datasets \citep{samek2017explainable, chen2020ai}. Regulatory bodies, including the U.S. Food and Drug Administration and the European Union’s AI Act, underscore the necessity for explainable systems in high-risk medical applications \citep{european2021ai}. Under Article 13 of the EU Artificial Intelligence Act, high-risk AI systems must be designed to ensure sufficient transparency. As stated in the regulation:

\begin{quote}
\textbf{Article 13: Transparency and provision of information to users}\\[1ex]
\textit{1. High-risk AI systems shall be designed and developed in such a way to ensure that their operation is sufficiently transparent to enable users to interpret the system’s output and use it appropriately. An appropriate type and degree of transparency shall be ensured to achieve compliance with the relevant obligations of the user and the provider set out in Chapter 3 of this Title.}
\end{quote}

Techniques such as Gradient-weighted Class Activation Mapping (Grad-CAM) for visualizing salient image regions \cite{selvaraju2017grad}, SHapley Additive exPlanations (SHAP) for quantifying feature contributions \cite{lundberg2017unified}, and attention mechanisms embedded in deep learning architectures \cite{vaswani2017attention} have demonstrated significant efficacy in enhancing model interpretability. These approaches are pivotal for ensuring transparency and accountability in artificial intelligence (AI) applications, particularly in healthcare, where explainability supports ethical deployment and contributes to improved clinical decision-making and patient outcomes.

In this work, we introduce \textbf{HQCM-EBTC}, a novel hybrid quantum-classical neural network that integrates classical convolutional feature extraction, an advanced attention mechanism, and parameterized quantum circuits for brain tumor classification. We design the model to extract essential imaging features using conventional CNN layers, while a dual-attention mechanism refines these features by emphasizing clinically relevant tumor regions. Furthermore, by incorporating quantum processing layers, our model captures complex non-linear correlations that may remain hidden in purely classical architectures. This integration not only enhances classification accuracy but also bolsters interpretability---an essential attribute for clinical applications where practitioners need to understand the basis of automated decisions.

Our experimental results demonstrate that HQCM-EBTC improves feature representation and classification performance while providing interpretable attention maps that align closely with ground-truth tumor regions. These findings underscore the potential of quantum-enhanced deep learning to offer more reliable and transparent diagnostic support in medical image analysis.

The remainder of this paper is structured as follows: Section \ref{LITERATURE REVIEW} reviews existing QML-based approaches for brain tumor classification; Section \ref{METHODOLOGY} details the architecture of HQCM-EBTC; Section \ref{RES} presents the experimental results; and Section \ref{DES} discusses the key findings, limitations, and future research directions.

\section{LITERATURE REVIEW}\label{LITERATURE REVIEW}
Recent studies in quantum machine learning (QML) have demonstrated promising advances in medical image analysis for brain tumor detection and classification. Amin et al. \citep{amin2023detection} presented a multi-stage framework that first improves MRI image quality with a 32-layer denoising neural network, and then employs a seven-layer quantum convolutional neural network (J. Qnet) for the classification of healthy and abnormal MRI slices. Their work further integrates lesion localization using an ONNX-YOLOv2tiny model and refines segmentation with a 34-layer U-net, achieving high accuracies on both local datasets and the BRATS-2020 benchmark. In another study, Amin et al. \citep{amin2022new} proposed an ensemble approach that extracts deep features using InceptionV3 and then classifies tumor types with a quantum variational classifier, followed by segmentation via a dedicated network; this model attained detection scores exceeding 90\% on multiple benchmark datasets. Similarly, Ullah et al. \citep{ullah2024brain} developed a framework combining deep learning with a quantum theory-based marine predator optimization algorithm, addressing dataset imbalance through sparse autoencoders and achieving accuracy levels nearing 99.80\% on augmented datasets. Khan et al. \citep{khan2024brain} further advanced the field with a quantum convolutional neural network (QCNN) tailored for brain tumor diagnosis, which achieved an accuracy of 99.67\%, highlighting the transformative potential of quantum models in clinical diagnostics.

Additional contributions in this domain include Gencer et al. \citep{gencer2025hybrid}, who integrated a pre-trained EfficientNetB0 with a novel quantum genetic algorithm for robust multi-class brain tumor classification, achieving accuracies above 98\% on various datasets. Choudhuri et al. \citep{choudhuri2023brain} proposed a quantum-classical convolutional network (QCCNN) that effectively encoded MRI data into quantum states, achieving classification accuracies in the range of 97.5–98.72\% across different datasets. Rajamohana et al. \citep{rajamohana2025hybrid} explored a hybrid quantum graph neural network (QGNN) that combined classical graph convolutional techniques with quantum-inspired models, demonstrating improved performance in differentiating normal from tumor images. Finally, Ticku et al. \citep{ticku2025advancing} introduced a quantum convolutional neural network architecture, developed using PennyLane and TensorFlow Quantum, which not only matched classical models in accuracy but also significantly reduced training times, thereby illustrating the practical benefits of quantum-enhanced diagnostic systems.

Our study builds upon these promising developments by proposing \textbf{HQCM-EBTC}, a novel hybrid quantum-classical neural network that integrates classical convolutional feature extraction, a dual-attention mechanism, and parameterized quantum circuits implemented via the PennyLane \citep{bergholm2022pennylaneautomaticdifferentiationhybrid}. Unlike prior works that often focus solely on improving classification accuracy or segmentation quality, our approach emphasizes the interpretability of the model's decision-making process through attention maps that closely align with clinically relevant tumor regions. HQCM-EBTC achieves superior in-distribution classification performance and exhibits enhanced out-of-distribution generalizability, addressing both the computational and diagnostic challenges inherent in brain tumor imaging. In this way, our work not only contributes to the growing body of QML-based diagnostic tools but also offers a robust, interpretable, and efficient framework for brain tumor classification that may significantly impact clinical practice.

\section{METHODOLOGY} \label{METHODOLOGY}

\begin{figure*}[!htb]
    \centering
\includegraphics[width=1\linewidth]{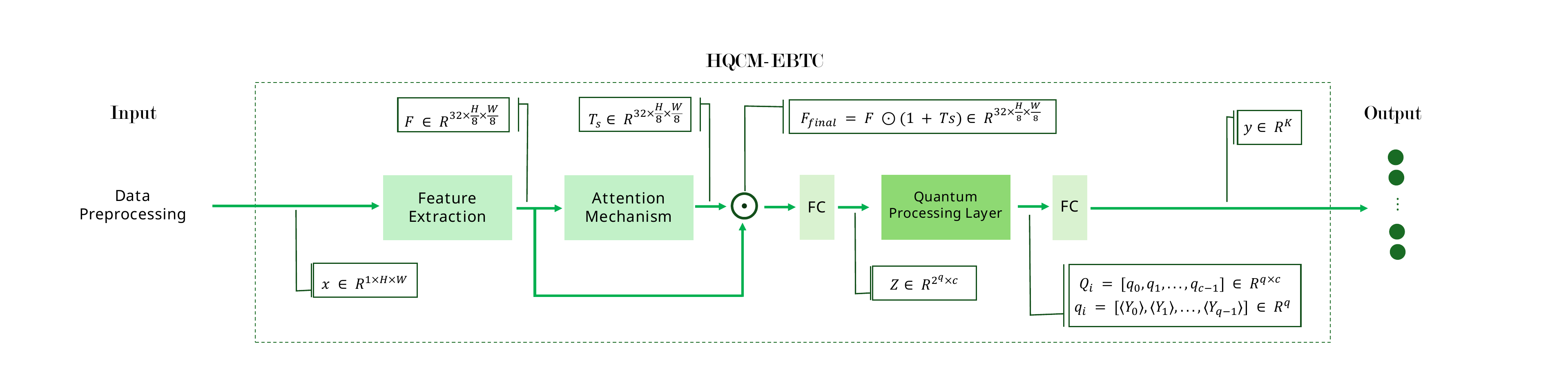}
    \caption{Overall architecture of the proposed \textbf{HQCM-EBTC} model. 
    Data preprocessing prepares input images \(\mathbf{x}\in\mathbb{R}^{1\times H \times W}\). 
    Classical convolutional blocks extract a feature tensor \(\mathbf{F}\in\mathbb{R}^{32\times \frac{H}{8}\times \frac{W}{8}}\), 
    which is refined by the attention mechanism to produce \(\mathbf{F}_{\text{final}}\). 
    This refined tensor is flattened and projected into \(\mathbf{z}\in\mathbb{R}^{2^q\times c}\), 
    then processed by the Quantum Processing Layer via amplitude embedding, entangling gates, 
    and measurement, yielding quantum outputs \(\mathbf{q}_i\in\mathbb{R}^q\). 
    Finally, these outputs are concatenated and passed to a fully connected layer for classification.}
    \label{fig:model}
    
\end{figure*}

Brain tumor classification from medical imaging posed significant challenges due to the complex and heterogeneous nature of tumor appearances, as well as the critical need for interpretability in clinical settings \citep{ranjbarzadeh2023brain} \citep{muller2022explainability}. Consequently, to address these challenges, we developed \textbf{HQCM-EBTC} (Figure \ref{fig:model}), a hybrid quantum-classical neural network that integrates classical convolutional feature extraction (Figure \ref{fig:FE}), an advanced dual-attention mechanism (Figure \ref{fig:AM}), and parameterized quantum circuits (Figure \ref{fig:QP}). In this regard, the architecture was designed not only to achieve high classification accuracy but also to enhance interpretability by ensuring that the model focuses on domain-specific regions of interest, such as tumor areas.

The overarching goal of HQCM-EBTC was to learn an optimal mapping from raw input images to class predictions while preserving the spatial and contextual information essential for medical diagnosis. Formally, we defined our dataset as
\[
\mathcal{D} = \{ (\mathbf{x}_i, y_i) \}_{i=1}^{N},
\]
where each input image \(\mathbf{x}_i \in \mathbb{R}^{C \times H \times W}\) consisted of \(C\) channels and spatial dimensions \(H \times W\), and \(y_i\) represented the corresponding ground-truth class label. Therefore, our objective was to approximate an optimal mapping function:
\[
f_\theta: \mathbb{R}^{C \times H \times W} \to \mathbb{R}^K,
\]
where \(K\) is the number of classification categories and \(\theta\) denotes the set of trainable model parameters. In doing so, we optimized \(f_\theta\) to minimize classification error while preserving attention consistency with critical tumor regions, thereby ensuring that the model's decisions remained both accurate and clinically interpretable.

In the following subsections, we describe the key components of HQCM-EBTC, detailing the methods employed to extract, refine, and process features from MRI images.

\subsection{Feature Extraction}
\begin{figure*}[!htb]
    \centering
    \includegraphics[width=1\linewidth]{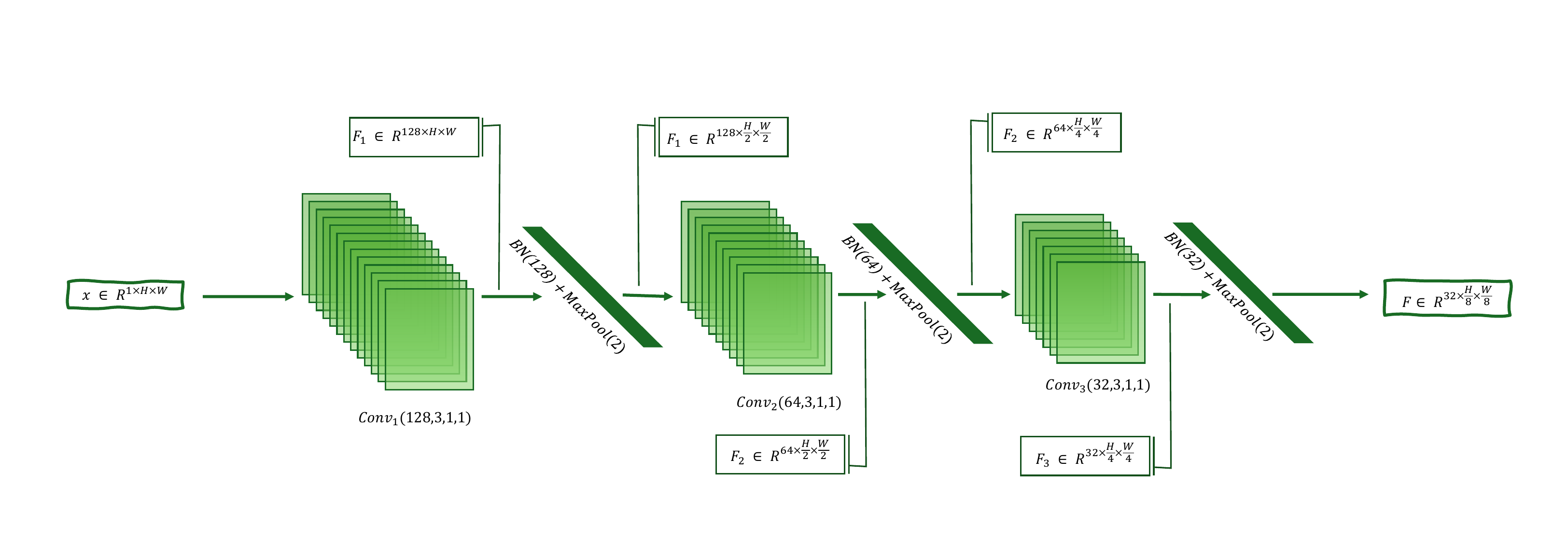}
    \caption{Illustration of the classical feature extraction pipeline. The input image (left) was successively processed by multiple convolutional blocks (each comprising convolution, batch normalization, ReLU activation, and max-pooling) to produce a progressively more abstract yet compact representation (right). The final feature maps served as the foundation for subsequent attention and quantum processing layers.}
    \label{fig:FE}
\end{figure*}

Initially, we designed the classical feature extraction module (Figure \ref{fig:FE}), which was responsible for extracting hierarchical representations from the input images. Given an input image 
\[
\mathbf{x} \in \mathbb{R}^{1 \times H \times W},
\]
where the single channel denoted a grayscale image, we employed three convolutional blocks to gradually extract and refine features at multiple levels of abstraction.

In the first convolutional block, we applied 128 filters with a \(3 \times 3\) kernel, stride 1, and padding 1 to the input image. Specifically, this operation was formulated as:
\[
\mathbf{y}_1 = \text{ReLU}\Bigl(\text{BN}\Bigl(\mathbf{W}_1 * \mathbf{x} + \mathbf{b}_1\Bigr)\Bigr),
\]
where \(\mathbf{W}_1 \in \mathbb{R}^{128 \times 1 \times 3 \times 3}\) and \(\mathbf{b}_1 \in \mathbb{R}^{128}\) were learnable parameters. Here, the convolution operator \(*\) slid each filter across the image to produce 128 feature maps that captured low-level patterns such as edges and textures. Moreover, batch normalization (BN) was applied immediately after the convolution to stabilize the learning process by normalizing activations, and subsequently, the ReLU activation was employed to introduce non-linearity that enabled the network to model complex functions.

Subsequently, additional convolutional blocks were employed to further refine these features. Each block incorporated extra convolutional layers to capture higher-level, more abstract representations and \(2 \times 2\) max-pooling operations to downsample the feature maps. In doing so, the network reduced spatial dimensions while capturing translation-invariant features and lowering computational complexity. Ultimately, after processing through all three convolutional blocks, the network produced a rich feature tensor 
\[
\mathbf{F} \in \mathbb{R}^{32 \times \frac{H}{8} \times \frac{W}{8}},
\]
which represented a compressed yet informative summary of the original image. Thus, this hierarchical extraction of features formed the robust foundation for subsequent processing stages.

\subsection{Attention Mechanism}
\begin{figure*}[!htb]
    \centering
    \includegraphics[width=1\linewidth]{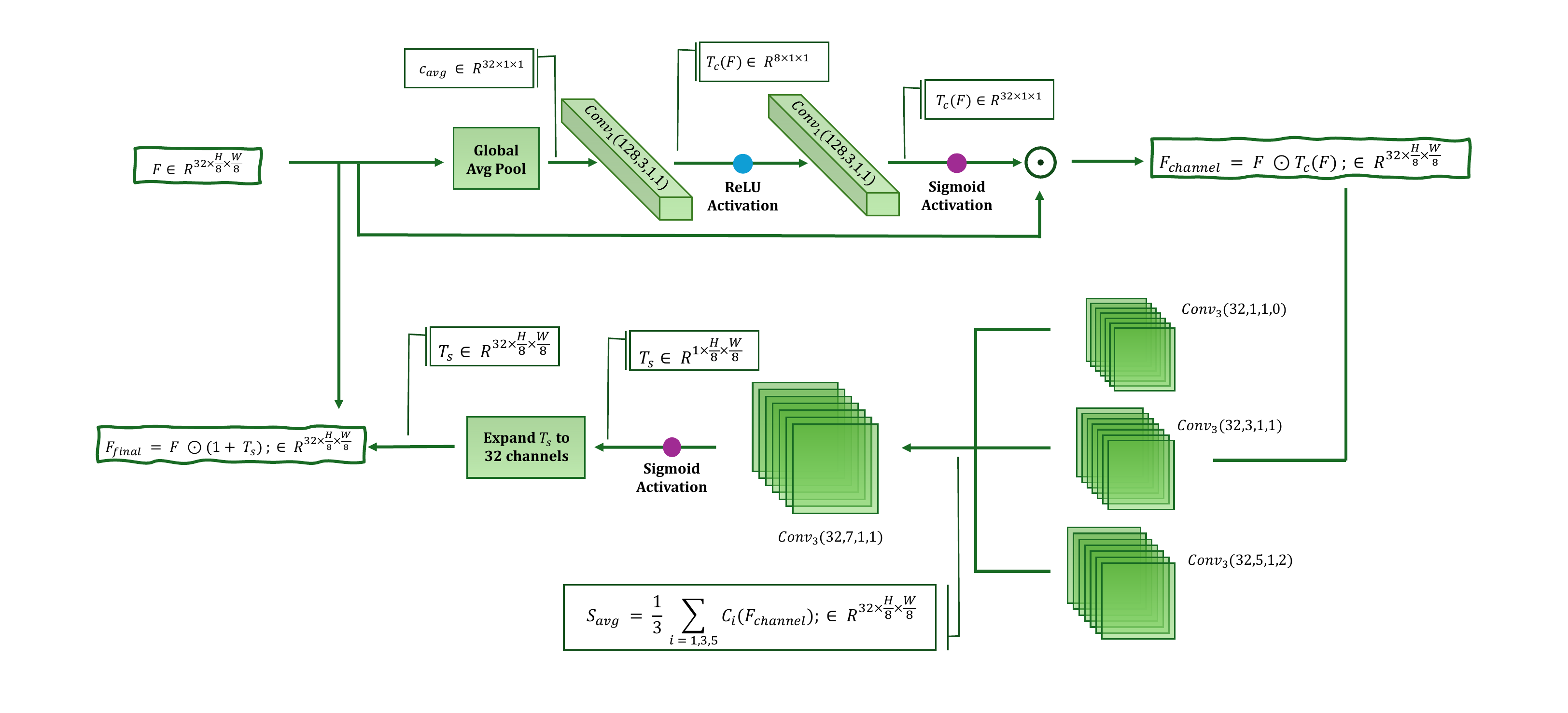}
    \caption{Visualization of the dual-attention module. The top branch (channel attention) aggregated global context via average pooling and learned channel weights through successive \(1\times1\) convolutions. These weights modulated the input feature maps to emphasize important channels. In parallel, the bottom branch (spatial attention) applied multi-scale convolutions and fused their outputs to generate a spatial attention map, which was then broadcast across all channels. The final attention-modulated feature maps combined both channel and spatial cues to highlight the most discriminative regions.}
    \label{fig:AM}
\end{figure*}

Inspired by the human visual system, which selectively focuses on salient regions \citep{Corbetta2002, Larochelle2010}, we designed an attention mechanism (Figure~\ref{fig:AM}) to refine feature representations and improve the interpretability of the network's predictions. This mechanism is based on the Convolutional Block Attention Module (CBAM) \citep{woo2018cbamconvolutionalblockattention}, which has demonstrated its effectiveness in enhancing the performance of convolutional neural networks.

Two attention modules, channel and spatial, were implemented to capture complementary attention, determining \textbf{what} channel contains the most relevant features and \textbf{where} the model should focus within the spatial domain. These modules are arranged sequentially, consistent with the findings of \citep{woo2018cbamconvolutionalblockattention}, which showed that this configuration allows the model to first attend to the most informative channels before refining the spatial localization of salient regions. Thus, our approach leverages CBAM's ability to adaptively weight features across channels and space, thereby enhancing the accuracy and understanding of the model's decisions.

\textbf{Channel attention:}
First, we computed a global context by applying a \emph{global average pooling} operation on the feature tensor \(\mathbf{F} \in \mathbb{R}^{C \times H' \times W'}\) produced by the feature extraction module. In this manner, the operation computed a compact descriptor for each channel by averaging over its spatial dimensions:
\[
\mathbf{c}_{\text{avg}} = \frac{1}{H' \times W'} \sum_{h=1}^{H'} \sum_{w=1}^{W'} \mathbf{F}(:,h,w),
\]
thus resulting in a vector \(\mathbf{c}_{\text{avg}} \in \mathbb{R}^{C}\) that captured the overall importance of each channel while reducing noise and redundancy.

Next, we passed the aggregated vector \(\mathbf{c}_{\text{avg}}\) through two successive \(1 \times 1\) convolutional layers with a reduction ratio, interleaved with a ReLU activation. This operation was represented as:
\[
\mathbf{T}_c(\mathbf{F}) = \sigma\left( \mathcal{C}_2\Bigl(\text{ReLU}\bigl(\mathcal{C}_1(\mathbf{c}_{\text{avg}})\bigr)\Bigr) \right),
\]
where \(\mathcal{C}_1\) and \(\mathcal{C}_2\) were convolution operations that learned non-linear interactions between channels, and \(\sigma\) was the sigmoid function scaling the output to the range \([0, 1]\). Consequently, the resulting channel attention weights \(\mathbf{T}_c(\mathbf{F}) \in \mathbb{R}^{C}\) indicated the relative importance of each channel. Thereafter, we applied these weights to the original feature tensor via element-wise multiplication:
\[
\mathbf{F}_{\text{channel}} = \mathbf{F} \odot \mathbf{T}_c(\mathbf{F}),
\]
which, in turn, enhanced the channels most relevant to the task.

\textbf{Spatial attention:}
Thereafter, we employed a spatial attention module to pinpoint and emphasize the regions within the feature maps that were most critical. To this end, and in order to capture spatial information at multiple scales, we processed the channel-refined features \(\mathbf{F}_{\text{channel}}\) using parallel convolutions with different kernel sizes (e.g., \(1 \times 1\), \(3 \times 3\), and \(5 \times 5\)). The outputs from these convolutions were then averaged:
\[
\mathbf{S}_{\text{avg}} = \frac{1}{3}\sum_{k \in \{1,3,5\}} \mathcal{C}_k\bigl(\mathbf{F}_{\text{channel}}\bigr)
\]

which effectively captured spatial patterns at various resolutions. Subsequently, we passed the fused spatial features \(\mathbf{S}_{\text{avg}}\) through a larger convolution (e.g., \(7 \times 7\)) followed by a sigmoid activation to generate the spatial attention map:
\[
\mathbf{T}_s = \sigma\left(\mathcal{C}_7(\mathbf{S}_{\text{avg}})\right),
\]
where \(\mathbf{T}_s \in \mathbb{R}^{1 \times H' \times W'}\) provided a weight for each spatial location. Finally, this map was broadcast across the channels and combined with the original features in a residual fashion:
\[
\mathbf{F}_{\text{final}} = \mathbf{F} \odot \Bigl(1 + \mathbf{T}_s\Bigr).
\]
In this way, adding \(1\) preserved the original feature intensities while amplifying the salient regions.

\subsection{Quantum Processing Layer}
\begin{figure*}[!htb]
    \centering
    \includegraphics[width=0.9\linewidth]{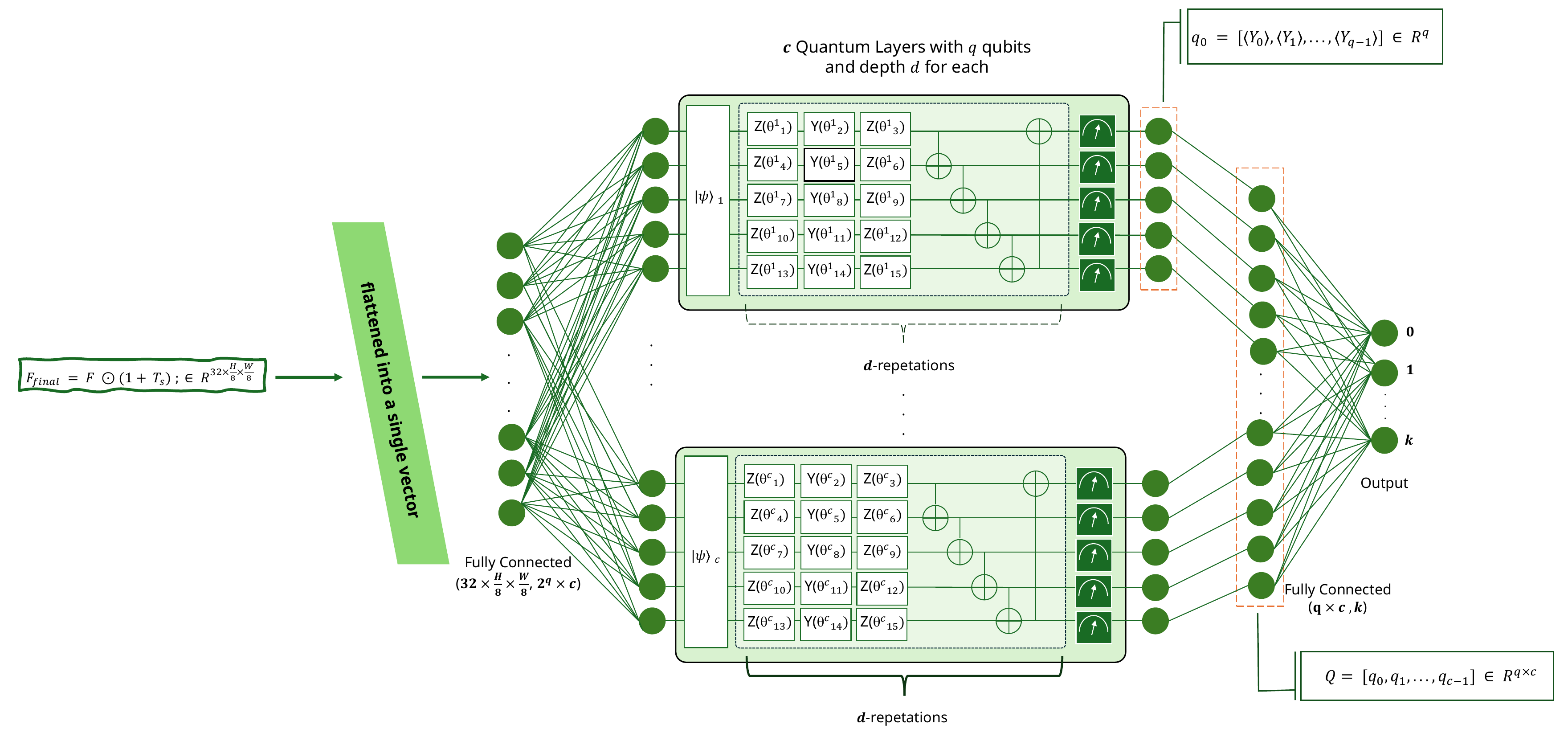}
    \caption{Schematic of the quantum processing layer. The classical feature vector was split and amplitude-embedded into \(c\) parallel quantum circuits, each with \(q\) qubits. Within each circuit, \(d\) layers of parameterized single-qubit rotations (\(R_z\), \(R_y\)) and controlled-NOT gates introduced entanglement and learnable transformations. The final measurements (shown as green measurement blocks) yielded expectation values that were concatenated across all circuits and passed to a classical fully connected layer for classification.}
    \label{fig:QP}
\end{figure*}

Subsequently, we integrated quantum processing (Figure \ref{fig:QP}) to capture complex non-linear correlations in the data by exploiting quantum phenomena such as superposition and entanglement. Once the attention mechanism refined the classical features, we flattened the resulting tensor 
\[
\mathbf{F}_{\text{final}} \in \mathbb{R}^{32 \times \frac{H}{8} \times \frac{W}{8}},
\]
and projected it into a vector 
\[
\mathbf{z} \in \mathbb{R}^{2^q \cdot c},
\]
where \(q\) represents the number of qubits per quantum circuit and \(c\) denotes the number of parallel quantum circuits. In this manner, the projection transformed the spatially distributed classical features into a format suitable for quantum encoding, thereby bridging the gap between classical image representations and quantum states.

Furthermore, each of the \(c\) parallel quantum circuits processed a sub-vector of \(\mathbf{z}\) via \textbf{amplitude embedding}. In this encoding scheme, we normalized the classical data and mapped it onto the amplitudes of a quantum state. Amplitude encoding was chosen because it efficiently maps normalized classical data onto the amplitudes of a quantum state, exploiting the full capacity of the Hilbert space. This method allows us to represent a large amount of information with a relatively small number of qubits—since \(n\) qubits can represent \(2^n\) amplitudes—thus providing a compact representation of high-dimensional data. In contrast, angle encoding (which uses rotation angles) or basis encoding (which directly maps data to computational basis states) often require more qubits or fail to capture the continuous nature of the data as effectively \citep{biamonte2017quantum, schuld2019quantum}.

\[
|\psi\rangle = \sum_{i=1}^{2^q} z_i \, |i\rangle,
\]
where \(z_i\) are the normalized amplitudes and \(|i\rangle\) denote the computational basis states of a \(q\)-qubit system. After that, the quantum circuit applied a sequence of strongly entangling gates, parameterized by a set of angles \(\theta\) and collectively represented by a unitary operator \(U(\theta)\). Specifically, the state transformation was defined as:
\[
|\psi(\theta)\rangle = U(\theta) |\psi\rangle.
\]
Here, we implemented \(U(\theta)\) as a series of layers, each composed of parameterized rotation gates and controlled-NOT (CNOT) operations. For instance, a typical rotation block was defined as:
\[
R_{\text{rot}}(\theta) = R_z(\theta_1) \, R_y(\theta_2) \, R_z(\theta_3),
\]
with
\[
R_z(\theta) = \begin{bmatrix} e^{-i\theta/2} & 0 \\ 0 & e^{i\theta/2} \end{bmatrix}
\]
and 
\[
R_y(\theta) = \begin{bmatrix} \cos(\theta/2) & -\sin(\theta/2) \\ \sin(\theta/2) & \cos(\theta/2) \end{bmatrix}.
\]
Subsequently, entanglement between qubits was introduced via CNOT gates, defined as:
\[
\text{CNOT} = \begin{bmatrix}
1 & 0 & 0 & 0 \\
0 & 1 & 0 & 0 \\
0 & 0 & 0 & 1 \\
0 & 0 & 1 & 0
\end{bmatrix}.
\]
By stacking \(d\) layers of parameterized rotation gates and CNOT operations, the circuit acquired the expressivity necessary to capture intricate patterns in the data.

After evolving through \(U(\theta)\), we measured each qubit to extract classical information from the quantum state. Specifically, we measured the expectation values of the Pauli-\(Y\) operators. For each qubit \(j\) in a circuit, the measurement was defined as:
\[
\langle Y_j \rangle = \langle \psi(\theta) | Y_j | \psi(\theta) \rangle,
\]
where \(Y_j\) denotes the Pauli-\(Y\) operator acting on qubit \(j\). Consequently, the measurement outcomes for each circuit were organized into a vector:
\[
\mathbf{q}_i = \left[\langle Y_0 \rangle, \langle Y_1 \rangle, \dots, \langle Y_{q-1} \rangle\right] \in \mathbb{R}^{q}.
\]
These measurement results capture the quantum circuit’s response to the parameterized operations, reflecting the complex correlations encoded by the entangling gates.

Finally, we concatenated the outputs from all \(c\) parallel quantum circuits to form a comprehensive quantum feature vector. This vector was subsequently fed into a classical fully connected layer:
\[
\mathbf{y} = \mathbf{W}_c \cdot \text{Concat}(\mathbf{q}_1, \dots, \mathbf{q}_c) + \mathbf{b}_c,
\]
where \(\mathbf{W}_c\) and \(\mathbf{b}_c\) are the weights and biases learned during training. In this way, the final layer effectively bridges the quantum and classical domains, producing the ultimate classification output.

\subsection{Training and Loss Function}
Subsequently, we trained the network using a composite loss function that integrated classification accuracy with attention consistency, thereby addressing the dual objectives of achieving high predictive performance and ensuring that the learned attention maps aligned with clinically relevant regions. Thus, the total loss was expressed as:
\[
\mathcal{L}_{\text{total}} = \alpha \, \mathcal{L}_{\text{class}} + \beta \, \mathcal{L}_{\text{attn}},
\]

Moreover, we defined the classification loss \(\mathcal{L}_{\text{class}}\) using the standard cross-entropy loss:
\[
\mathcal{L}_{\text{class}} = - \sum_{i=1}^{N} w_i \, y_i \log(p_i),
\]
where \(w_i\) denoted the class weight for the \(i\)-th class, \(y_i\) was the true label, and \(p_i\) was the predicted probability. It is worth noting that cross-entropy loss has been widely adopted in classification tasks for its theoretical grounding in information theory and its effectiveness in penalizing overly confident yet incorrect predictions \citep{mao2023crossentropylossfunctionstheoretical}.

In addition, to ensure that the network's attention maps corresponded closely with ground-truth tumor masks, we incorporated an attention consistency loss \(\mathcal{L}_{\text{attn}}\). This loss combined binary cross-entropy (BCE) and Dice losses \citep{azad2023lossfunctionserasemantic}:
\[
\mathcal{L}_{\text{attn}} = \zeta \, \mathcal{L}_{\text{BCE}} + \gamma \, \mathcal{L}_{\text{Dice}},
\]
with \(\zeta\) and \(\gamma\) selected to emphasize spatial overlap. Specifically, the BCE component was defined as:
\[
\mathcal{L}_{\text{BCE}} = -\frac{1}{B} \sum_{i=1}^B \left[ \mathbf{T}_i \log \mathbf{A}_i + (1 - \mathbf{T}_i) \log(1 - \mathbf{A}_i) \right],
\]
which quantified the pixel-wise discrepancy between the predicted attention map \(\mathbf{A}\) and the binary tumor mask \(\mathbf{T}\). Furthermore, in view of the class imbalance—where tumor regions occupy a small fraction of the image—we incorporated the Dice loss:
\[
\mathcal{L}_{\text{Dice}} = 1 - \frac{2 \sum_{i=1}^B \mathbf{T}_i \mathbf{A}_i + \epsilon}{\sum_{i=1}^B \mathbf{T}_i + \sum_{i=1}^B \mathbf{A}_i + \epsilon},
\]
where \(\epsilon\) was a small constant to prevent division by zero. In summary, the combination of BCE and Dice loss provided a balanced approach that captured fine details while ensuring overall region consistency.

\section{RESULTS} \label{RES}
In this section, we present a detailed analysis of the results obtained from our proposed HQCM-EBTC model and compare its performance to a classical counterpart in the task of brain tumor classification. We begin by describing the dataset construction and preprocessing steps, followed by an overview of the hyperparameter settings used during training. We then analyze the training performance, highlighting key metrics such as loss and accuracy, and provide a quantitative evaluation of the model's performance. The section also includes an in-depth examination of t-SNE visualizations, confusion matrices, and attention map analysis, all of which underscore the advantages of the quantum-enhanced model in terms of classification accuracy, tumor localization, and overall generalization.

\subsection{Dataset Construction and Preprocessing}
\begin{figure*}[ht]
    \centering
    \includegraphics[width=0.8\linewidth]{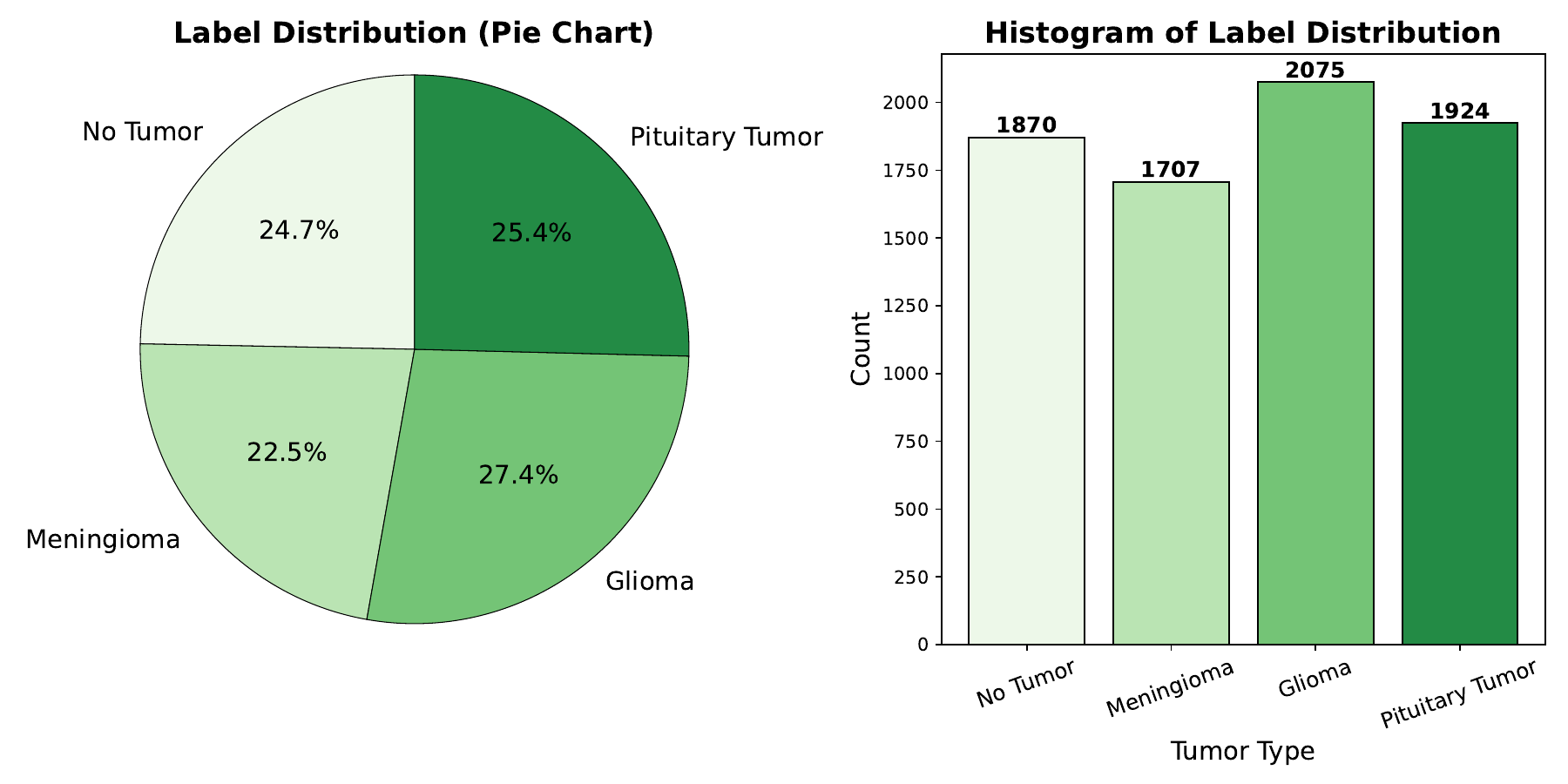}
    \caption{Combined data distribution in the integrated brain tumor dataset. The left panel shows a pie chart with the relative proportions of each class (Healthy Brain (0), Meningioma (1), Glioma (2), and Pituitary Tumor (3)), and the right panel displays a histogram of sample counts.}
    \label{fig:dataset_distribution}
\end{figure*}

\begin{figure}[H]
    \centering
    \includegraphics[width=0.8\linewidth]{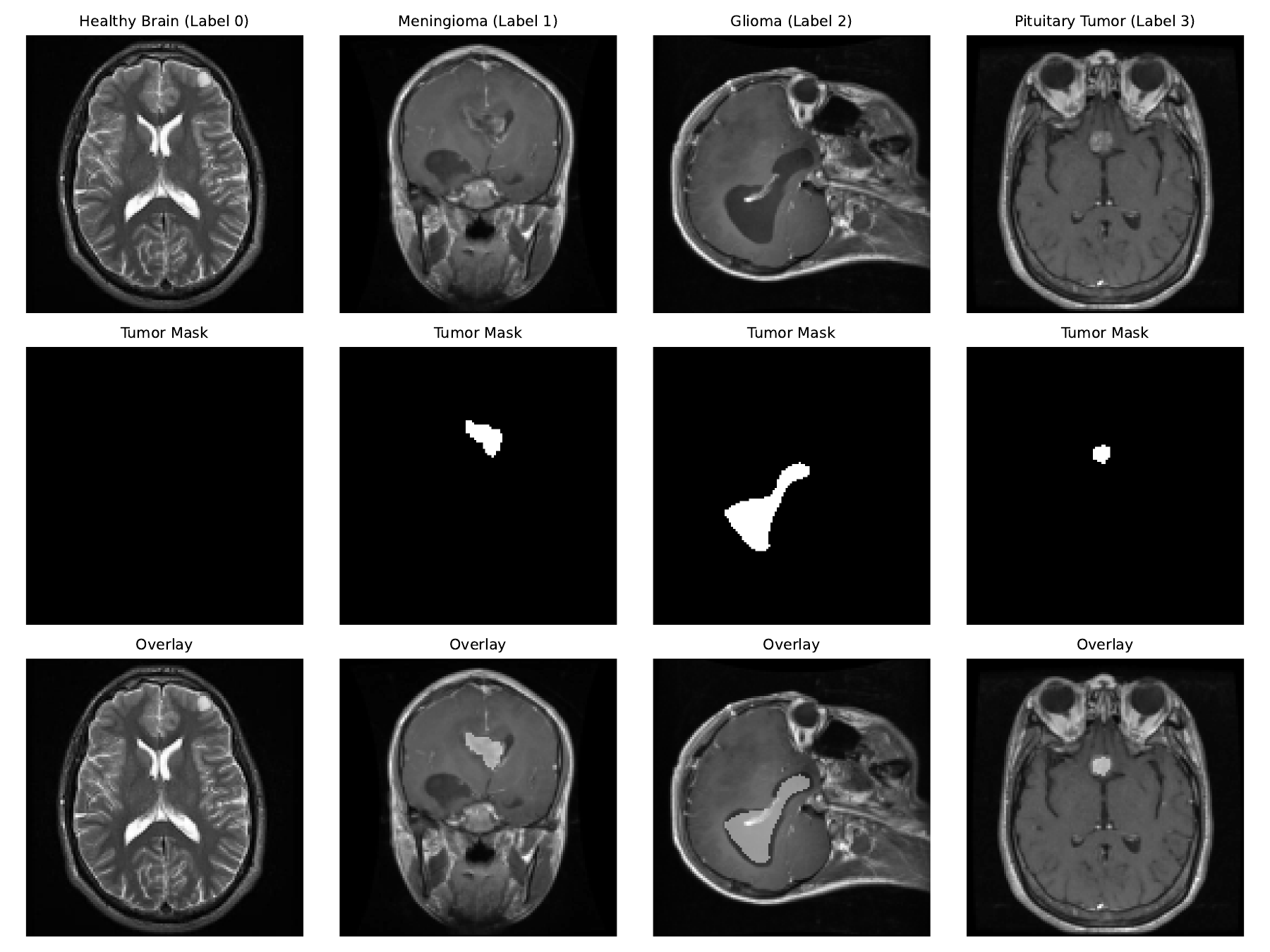}
    \caption{Sample images from the integrated dataset, illustrating normal brain structures and various tumor types.}
    \label{fig:dataset_samples}
\end{figure}

A brain tumor dataset of 7,576 images was constructed by merging data from three sources. Source 1 \citep{trainingdatapro_dicom_brain_dataset} contributed 275 images of normal brain structures. Source 2 \citep{jun2017brain} provided 3,064 T1-weighted MRI slices, including 708 meningioma, 1,426 glioma, and 930 pituitary tumor slices, which were converted into standard image formats. Additional images were obtained from Source 3 \citep{akter2024robust}. All images were resized to 
128×128 pixels and normalized using min-max scaling. The dataset was split into 70\% for training, 15\% for evaluation, and 15\% for testing, with a batch size of 64 during training. Data augmentation techniques, such as random rotations, and horizontal and vertical flips, were applied to improve generalization. The dataset is structured for deep learning applications, ensuring proper image and mask processing for accurate classification and segmentation. Figure~\ref{fig:dataset_distribution} shows the class distribution, and Figure~\ref{fig:dataset_samples} presents sample images.

\subsection{Hyperparameter Settings}
HQCM-EBTC was trained for 40 epochs using AdamW (learning rate 0.01, weight decay $1 \times 10^{-4}$). Gradient clipping (max norm 1.0) and ReduceLROnPlateau (factor 0.5, patience 3) were used; early stopping was applied after 5 epochs. The quantum layer, implemented via PennyLane, utilized 5 qubits per circuit, depth 2, and 5 parallel circuits. A composite loss (cross-entropy + Dice-based attention consistency) with weights $\alpha = 1$, $\beta = 1$, $\zeta = 0.3$, $\gamma = 0.7$ was employed. Key hyperparameters are summarized in Table~\ref{tab:training_params}.
\subsection{Training Performance Analysis}
\begin{figure*}[!htb]
    \centering  
    \includegraphics[width=1\linewidth]{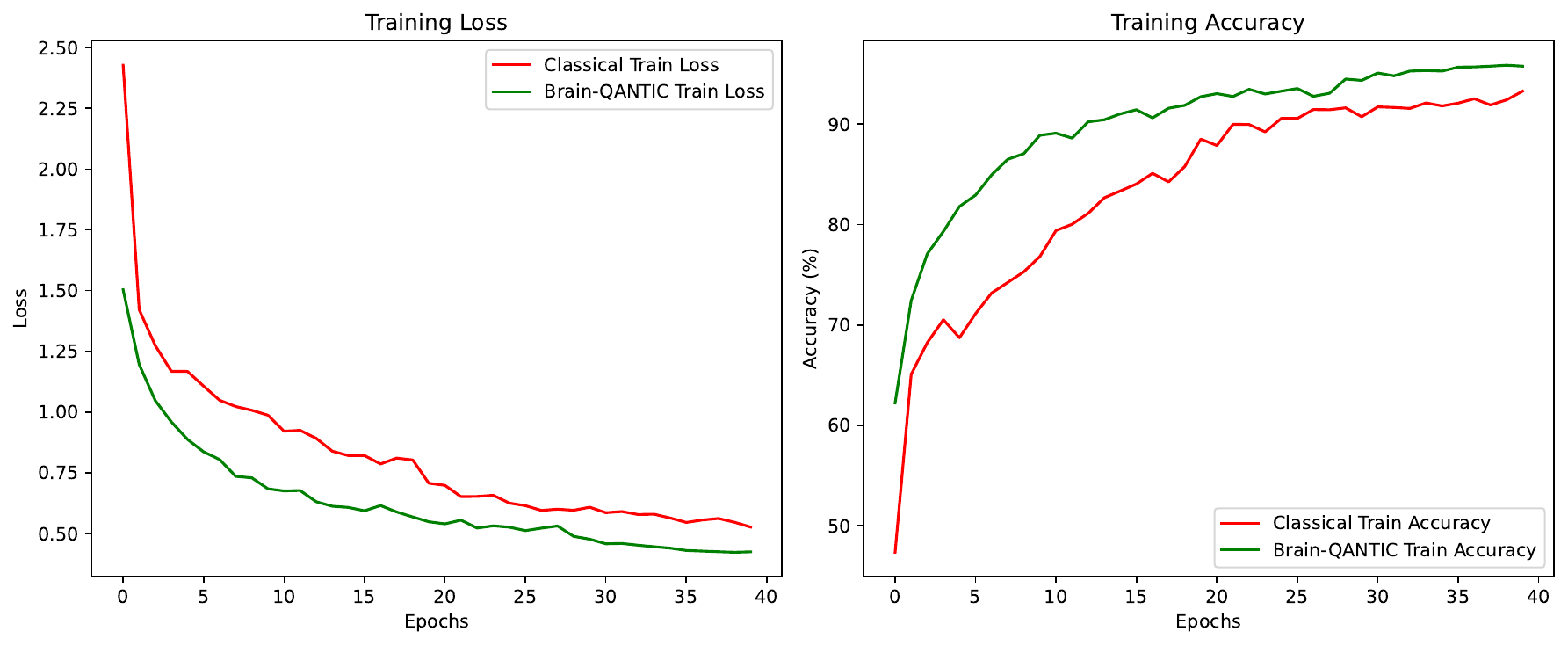}
    \caption{Training and validation loss (left) and accuracy (right) over epochs. The solid lines represent training metrics, while the dashed lines denote validation metrics.}
    \label{fig:TR}
\end{figure*}

Figure \ref{fig:TR} shows the training loss and the training accuracy for both the HQCM-EBTC and Classical models. The HQCM-EBTC model exhibits a lower initial training loss (1.50 vs. 2.43) and higher final accuracy (95.87\% vs. 93.29\%) compared to the classic model. A faster convergence rate is observed in HQCM-EBTC, with a consistently lower loss and higher accuracy throughout the training, indicating better optimization and generalization.

\subsection{Quantitative Performance Evaluation}

\begin{table*}[ht]
    \centering
    \caption{Performance Metrics of HQCM-EBTC and Classical Counterpart on the Test Dataset}
    \label{tab:classification_report}
    \resizebox{\textwidth}{!}{
    \begin{tabular}{lcccccccccccc}
        \hline
        & \multicolumn{3}{c}{\textbf{HQCM-EBTC}} & & \multicolumn{3}{c}{\textbf{Classical Counterpart}} \\
        
        \textbf{Class} & \textbf{Precision} & \textbf{Recall} & \textbf{F1-score} & & \textbf{Precision} & \textbf{Recall} & \textbf{F1-score} \\
        \hline
        0 & 0.99 & 0.97 & 0.98 & & 0.91 & 0.98 & 0.94 \\
        1 & 0.96 & 0.91 & 0.93 & & 0.80 & 0.88 & 0.84 \\
        2 & 0.94 & 0.98 & 0.96 & & 0.99 & 0.69 & 0.82 \\
        3 & 0.96 & 0.99 & 0.97 & & 0.84 & 0.99 & 0.91 \\
        \hline
        \textbf{Overall Accuracy} & \multicolumn{3}{c}{96.48\%} & & \multicolumn{3}{c}{86.72\%} \\
        \textbf{Macro Avg} & 0.96 & 0.96 & 0.96 & & 0.88 & 0.88 & 0.88 \\
        \textbf{Weighted Avg} & 0.96 & 0.96 & 0.96 & & 0.89 & 0.88 & 0.87 \\
        \hline
    \end{tabular}
    }
\end{table*}

The performance evaluation of HQCM-EBTC and its Classical Counterpart (Table \ref{tab:classification_report}) reveals a significant advantage in favor of the quantum-enhanced model. HQCM-EBTC achieves an overall accuracy of 96.48\%, outperforming the Classical Counterpart, which attains an accuracy of 86.72\%. 

Examining precision, recall, and F1 score in all classes, HQCM-EBTC consistently demonstrates superior values. Notably, in class 2, the Classical Counterpart struggles with a recall of only 0.69, whereas HQCM-EBTC achieves 0.98. This indicates that the quantum-enhanced model is significantly more effective at correctly identifying samples from this class, thereby reducing misclassification rates. The macro and weighted averages further reinforce this trend, highlighting the robustness of HQCM-EBTC across different classes.

\begin{figure*}[ht]
    \centering
    \subfloat[t-SNE projection for Quantum Processing Layer]{%
        \includegraphics[width=0.45\linewidth]{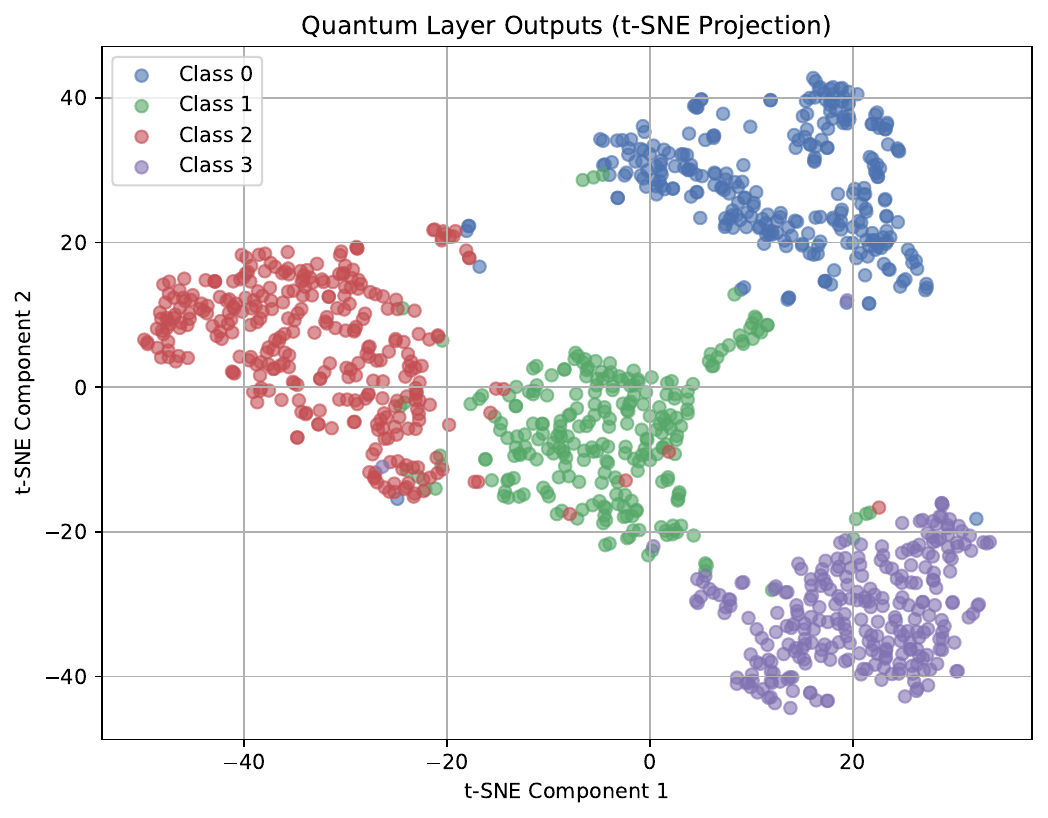}%
        \label{fig:QPL_TSNE}
    }
    \qquad
    \subfloat[t-SNE projection for Fully Connected Layer]{%
        \includegraphics[width=0.45\linewidth]{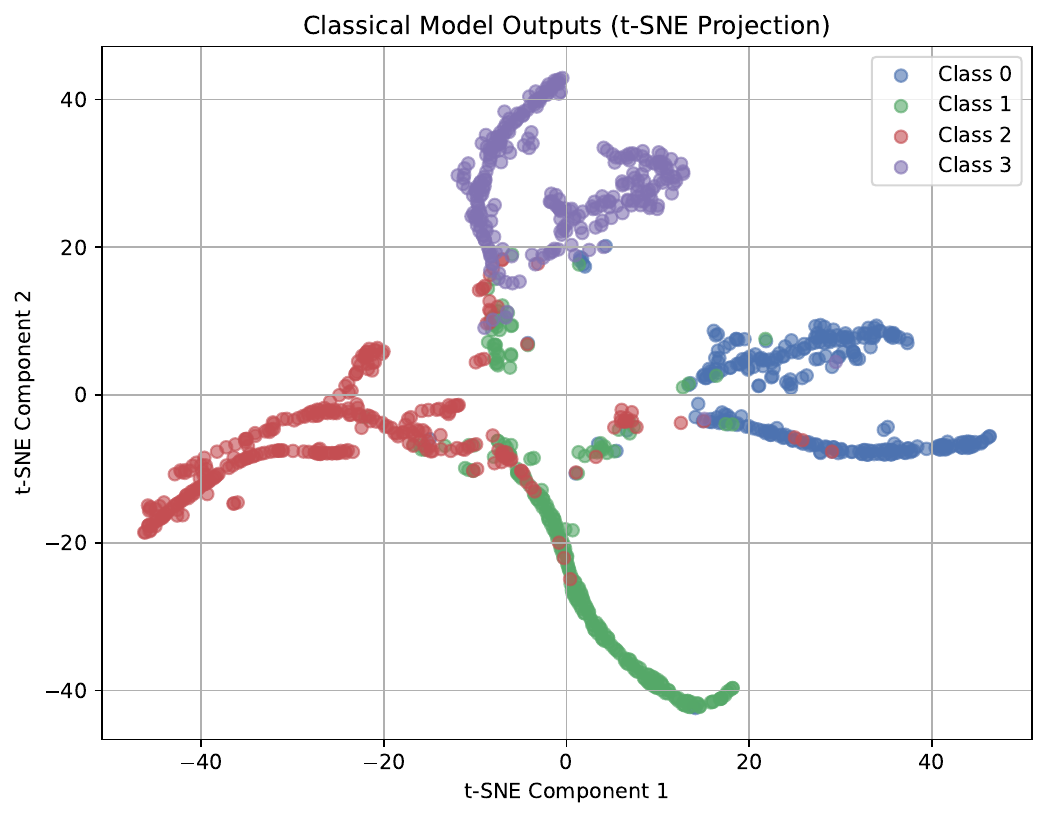}%
        \label{fig:FC_layer_TSNE}
    }
    \caption{Comparison of \textit{t}-SNE projections for the Quantum Processing Layer and the Fully Connected Layer. The visualization illustrates the two-dimensional embeddings of test samples, with colors indicating distinct classes. A perplexity of 30 and a fixed random initialization seed of 42 were used to ensure consistency across the projections.}

    \label{fig:combined_TSNE}
\end{figure*}

These quantitative results align with the t-SNE visualizations presented in Figures~\ref{fig:combined_TSNE}, which provide a qualitative perspective on the model's enhanced classification capabilities. The t-SNE projection \citep{van2014visualizing} of the Quantum Processing Layer (QPL) (Figure~\ref{fig:QPL_TSNE}) demonstrates more compact and well-separated clusters, particularly for Class 0 and Class 2. In contrast, the fully connected layer's projection (Figure~\ref{fig:FC_layer_TSNE}) exhibits more elongated and overlapping clusters, suggesting that its learned feature representations are less distinct. The improved separability in the QPL's feature space translates into clearer decision boundaries, which in turn contribute to HQCM-EBTC's superior classification accuracy. 

By leveraging quantum-enhanced feature extraction, HQCM-EBTC generates a more discriminative embedding space characterized by reduced class overlap and improved generalization. This observation not only underscores the inherent advantages of quantum processing in high-dimensional learning tasks but also reinforces its potential to deliver robust classification performance in complex scenarios.

\subsection{Confusion Matrix Analysis}
\begin{figure*}[ht]
    \centering
    \subfloat[Confusion matrix for HQCM-EBTC]{%
        \includegraphics[width=0.45\linewidth]{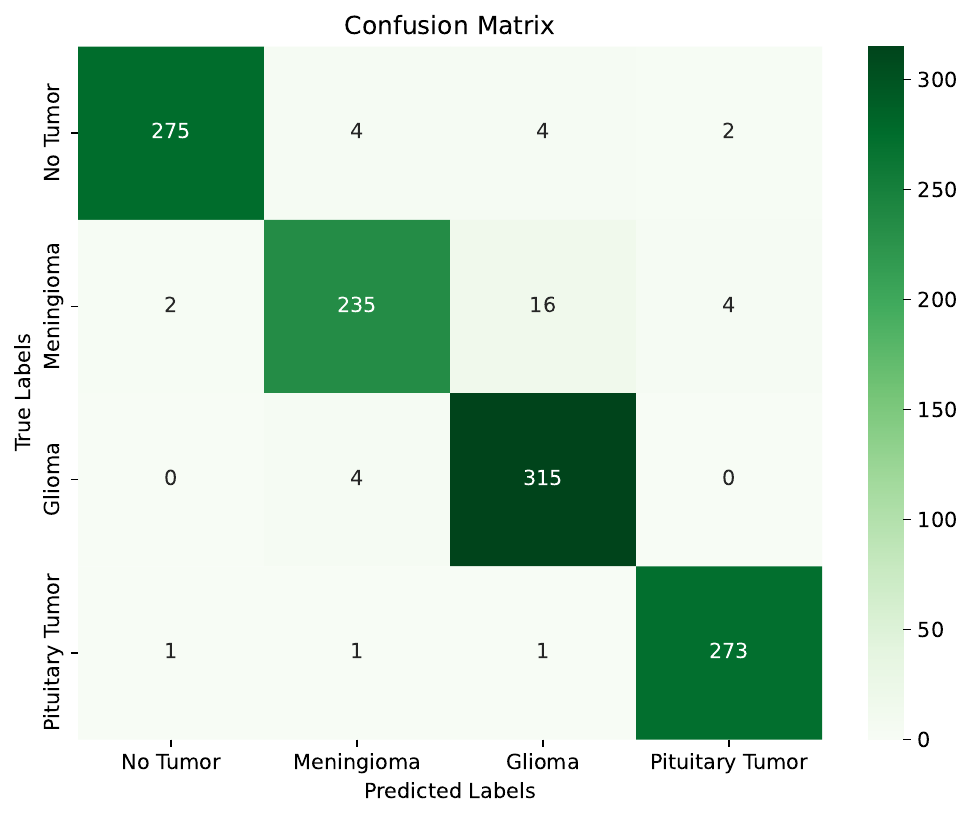}%
        \label{fig:CM_Q}
    }
    \qquad
    \subfloat[Confusion matrix for Classical Counterpart]{%
        \includegraphics[width=0.45\linewidth]{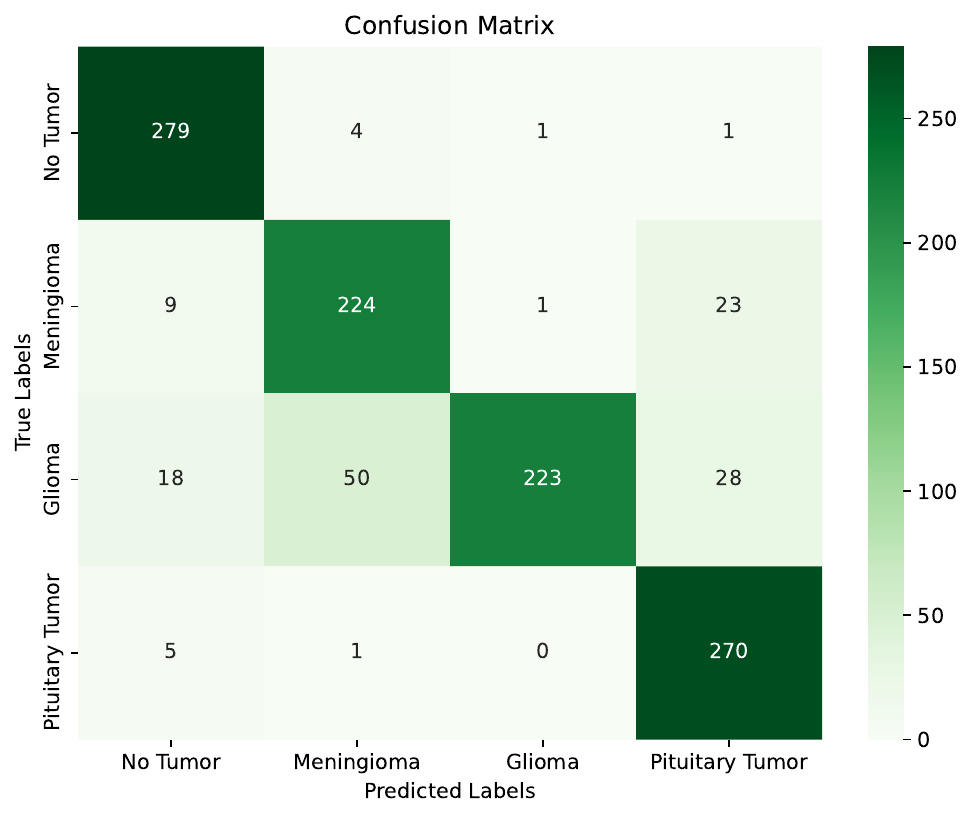}%
        \label{fig:CM_C}
    }
    \caption{Comparison of confusion matrices for HQCM-EBTC and its Classical Counterpart, illustrating the distribution of actual versus predicted labels.}
    \label{fig:combined_CM}
\end{figure*}

The confusion matrices (Figure \ref{fig:combined_CM}) further emphasize the enhanced classification performance of HQCM-EBTC. The Classical Counterpart exhibits notable misclassifications, particularly in Class 2 (Glioma), where a significant number of instances are misclassified as Class 1 (Meningioma) and Class 3 (Pituitary). In contrast, HQCM-EBTC demonstrates a lower misclassification rate, particularly in distinguishing Glioma from other tumor classes. This reduction in classification errors highlights the superior feature discrimination ability of HQCM-EBTC, contributing to its higher accuracy and reliability in medical image classification tasks.

\subsection{Attention Map Analysis}

\begin{figure*}[!htb]
    \centering
    \includegraphics[width=1\linewidth]{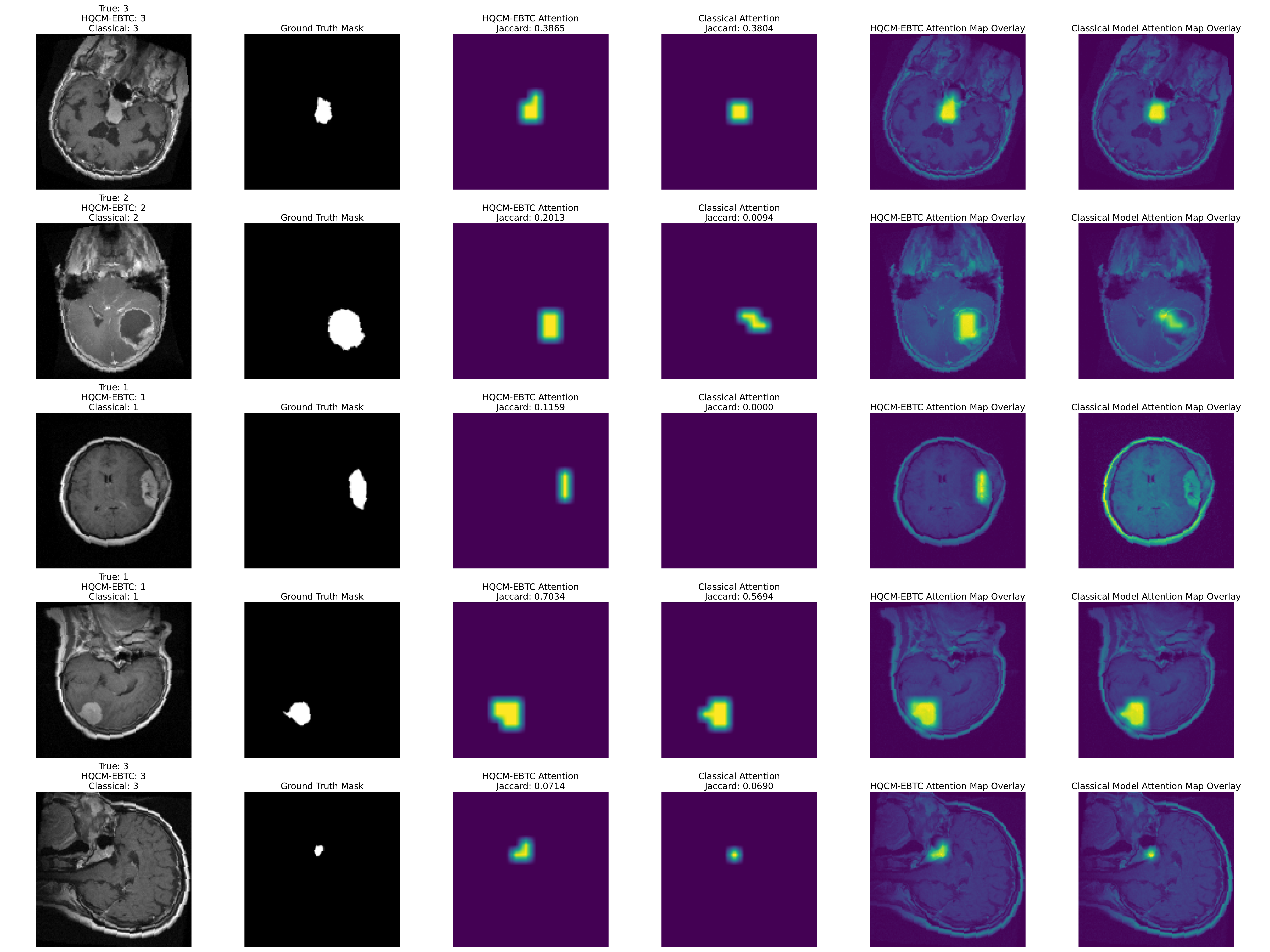}
    \caption{
        Comparison of attention maps generated by the quantum-enhanced model (HQCM-EBTC) and the classical model at \(\tau = 0.99\). Each row represents a test image, with six columns showing: (1) the input MRI image, (2) the ground-truth tumor mask, (3) the HQCM-EBTC attention map with its Jaccard Index, (4) the classical model’s attention map with its Jaccard Index, (5) the overlay of the HQCM-EBTC attention map on the input image, and (6) the overlay of the classical model’s attention map. The Jaccard Index (IoU) is computed at a threshold of \(\tau = 0.99\), highlighting the most confidently activated pixels. Higher Jaccard values indicate better alignment between the attention map and the ground-truth tumor region.
    }
    
    \label{fig:attention_maps2}
\end{figure*}

We compare the attention maps of the HQCM-EBTC and its classical counterpart model in Figure~\ref{fig:attention_maps2}. HQCM-EBTC produces more focused activations that closely align with the ground-truth masks, while the classical counterpart exhibits broader, less localized activations. For non-tumor images, HQCM-EBTC maintains low activation levels, reducing false positives.

To quantify the alignment, we compute the Jaccard Index (IoU) at three thresholds: \(\tau = 0.99\), \(\tau = 0.90\), and \(\tau = 0.75\), using the formula:

\[
J(A_{\tau}, M) = \frac{| A_{\tau} \cap M |}{| A_{\tau} \cup M |},
\]

where \(A_{\tau}\) is the binarized attention map at threshold \(\tau\) and \(M\) is the ground-truth mask. 

The results show that HQCM-EBTC outperforms the classical counterpart at \(\tau = 0.99\) (Jaccard Index: 0.43220 vs. 0.40219), with a statistically significant difference (\(p = 0.0017\)). At \(\tau = 0.90\), the difference remains significant (\(p = 0.0350\)), but no significant difference is found at \(\tau = 0.75\) (\(p = 0.1064\)).

These findings suggest that HQCM-EBTC is better at localizing tumor regions, especially at higher confidence thresholds. Figure~\ref{fig:attention_maps2} illustrates this by overlaying the attention maps on the input images, showing that HQCM-EBTC’s attention is more localized and aligned with the ground-truth tumors.

\begin{table*}[!ht]
    \centering
    \caption{Jaccard Index Comparison at Different Thresholds.}
    \label{tab:jaccard_comparison}
    \resizebox{\textwidth}{!}{
    \begin{tabular}{c|c|c|c|c}
        \hline
        \textbf{Threshold} (\(\tau\)) & \textbf{HQCM-EBTC} & \textbf{Classical Counterpart} & \textbf{Wilcoxon Test Statistic} & \textbf{p-value} \\  
        \hline
        \textbf{0.99}  & \textbf{0.43220} & 0.40219 & 914737.5 & \textbf{0.0017}  \\
        \textbf{0.90}  & \textbf{0.39353} & 0.36829 & 921261.5 & \textbf{0.0350}  \\
        \textbf{0.75}  & \textbf{0.37609} & 0.36664 & 950092.0 & 0.1064 \\
        \hline
    \end{tabular}
    }
\end{table*}

\subsection{Conclusion}

The evaluation of HQCM-EBTC in comparison to its classical counterpart reveals substantial improvements in both classification performance and tumor localization. HQCM-EBTC achieves an overall accuracy of 96.48\%, surpassing the classical model's accuracy of 86.72\%. This is further supported by higher precision, recall, and F1-scores across all tumor classes, particularly in Class 2 (Glioma), where the classical model shows significantly lower recall (0.69) compared to HQCM-EBTC's 0.98 (Table \ref{tab:classification_report}).

The t-SNE visualizations (Figure \ref{fig:combined_TSNE}) highlight the enhanced separability of the quantum model's feature space, with the Quantum Processing Layer (QPL) showing well-defined, compact clusters, contrasting with the more overlapping clusters of the classical counterpart's fully connected layer. This greater separability contributes to HQCM-EBTC’s superior classification accuracy by providing clearer decision boundaries.

Further analysis of the confusion matrices (Figure \ref{fig:combined_CM}) confirms that HQCM-EBTC outperforms the classical model, particularly in reducing misclassifications within Class 2 (Glioma). The attention map analysis (Figure \ref{fig:attention_maps2}) shows that HQCM-EBTC produces more localized and focused activations that align more closely with the ground-truth tumor masks, particularly at higher confidence thresholds, as evidenced by the Jaccard Index comparisons (Table \ref{tab:jaccard_comparison}). These results underscore the quantum model’s ability to better localize tumor regions, enhancing its clinical applicability.

\section{Discussion} \label{DES}

This study demonstrates the potential of the Quantum-enhanced Hybrid Classification Model (HQCM-EBTC) for medical image classification, particularly in tumor detection from MRI scans. Our model outperforms classical models, achieving a higher overall accuracy of 96.48\% compared to 86.72\% in the classical approach. This improvement is particularly evident in more challenging tumor types, such as Glioma, where HQCM-EBTC significantly enhances recall, capturing subtle tumor features that classical models often misclassify.

The improved performance is also supported by visualization methods like t-SNE and confusion matrices, which show better separability between tumor classes in the quantum-enhanced feature space. Additionally, attention map analysis reveals that HQCM-EBTC focuses more precisely on relevant regions of the image, which aligns better with tumor boundaries, further contributing to its superior classification results.

Despite these promising results, there are still challenges to overcome. Quantum circuit complexity and hardware limitations remain key obstacles. While the model performs well on the current dataset, scaling it for larger datasets or real-time clinical use requires further optimization. Additionally, the potential for overfitting to the current dataset calls for further validation on more diverse datasets to ensure generalizability.

Looking forward, future work should focus on optimizing quantum circuits for better efficiency, addressing the scalability issues, and exploring the integration of other data modalities like genetic or clinical data. Federated learning approaches could enable collaborative training while preserving privacy, and causal inference methods could enhance the model's interpretability, offering insights into underlying tumor development factors. 

In conclusion, the HQCM-EBTC model represents a significant advancement in applying quantum computing to medical image analysis. While challenges remain, its potential to improve diagnostic accuracy and interpretability in healthcare is promising, offering a glimpse of how quantum-enhanced models could transform the field of medical diagnostics.


\section*{Declarations}
\bigskip
\textbf{Author Statement:}
Marwan Ait Haddou conducted the research, designed the hybrid model, performed experiments, and wrote the manuscript. Mohamed Bennai supervised the project, provided critical guidance throughout, and contributed to the refinement of the manuscript. Both authors reviewed and approved the final version of the manuscript. The authors declare no competing interests, and the manuscript is not under consideration elsewhere.

\bigskip
\textbf{Funding}
No funding was received for conducting this study.

\bigskip
\textbf{Conflicts of Interest:} The authors declare no competing interests.

\bigskip
\textbf{Generative AI statement:}
 The authors declare that Generative AI was used in the creation of this manuscript. Generative AI was used for the refinement of English-language editing in a few sections.
\bigskip

\textbf{Availability of Data and Materials:} The datasets and numerical details necessary to replicate this work are available from the corresponding author upon reasonable request.
\newpage
\newpage
\bibliographystyle{plain}
\bibliography{sample}
\onecolumn
\newpage

\begin{table*}[h]
    \centering
    \caption{Quantum and Classical Model Training Configuration}
    \label{tab:training_params}
    \begin{tabular}{l c}
        \hline
        \textbf{Parameter} & \textbf{Value} \\
        \hline
        Optimizer & AdamW \\
        Learning Rate & 0.011 \\
        Weight Decay & \(1 \times 10^{-4}\) \\
        Gradient Clipping (Max Norm) & 1.0 \\
        LR Scheduler & ReduceLROnPlateau (factor=0.5, patience=3) \\
        Early Stopping & 5 epochs \\
        Maximum Epochs & 40 \\
        Batch Size & 64 \\
        \hline
        \textbf{Quantum Processing Layer} & \\
        \hline
        Qubits per Circuit & 5 \\
        Circuit Depth & 2 \\
        Parallel Circuits & 5 \\
        \hline
        \textbf{Loss Function Weights} & \\
        \hline
        Classification Loss (\(\mathcal{L}_{\text{class}}\)) Weight (\(\alpha\)) & 1 \\
        Attention Loss (\(\mathcal{L}_{\text{attn}}\)) Weight (\(\beta\)) & 1 \\
        Dice Loss Scaling (\(\zeta\)) & 0.3 \\
        BCE Loss Scaling (\(\gamma\)) & 0.7 \\
        \hline
        \textbf{Hardware and Training Time} & \\
        \hline
        GPU & Tesla T4 (Google Colab) \\ 
        \hline
    \end{tabular}
\end{table*}

\end{document}